\documentclass[conference]{IEEEtran}
\IEEEoverridecommandlockouts
\usepackage{cite}
\usepackage{amsmath,amssymb,amsfonts}
\usepackage{hyperref}
\usepackage{graphicx}
\usepackage{textcomp}
\usepackage{xcolor}    
\usepackage{subfig}
\usepackage{pdfpages}
\usepackage{caption}
\def\BibTeX{{\rm B\kern-.05em{\sc i\kern-.025em b}\kern-.08em
    T\kern-.1667em\lower.7ex\hbox{E}\kern-.125emX}}

\definecolor{hollywoodcerise}{rgb}{0.96, 0.0, 0.63}
\definecolor{somecolor}{rgb}{0., 0.96, 0.63}

\newcommand{\IF}{\mathrm{IF}}
\newcommand{\PAC}{\mathrm{PAC}}
\begin{document}

\title{Distribution and volume based scoring for Isolation Forests \\
}

\author{\IEEEauthorblockN{1\textsuperscript{st} Hichem Dhouib}
\IEEEauthorblockA{
\textit{Porsche Digital GmbH}\\
Berlin \\
hichem.dhouib@porsche.digital}
\and
\IEEEauthorblockN{2\textsuperscript{nd} Alissa Wilms}
\IEEEauthorblockA{\textit{Freie Universität Berlin} \\
\textit{Porsche Digital GmbH}\\
Berlin \\
alissa.wilms@porsche.digital}
\and
\IEEEauthorblockN{3\textsuperscript{rd} Paul Boes}
\IEEEauthorblockA{\textit{Porsche Digital GmbH} \\
Berlin \\
paul.boes@porsche.digital}}

\maketitle

\begin{abstract}
We make two contributions to the \emph{Isolation Forest} method for anomaly and outlier detection. The first contribution is an information-theoretically motivated generalisation of the score function that is used to aggregate the scores across random tree estimators. This generalisation allows one to take into account not just the ensemble average across trees but instead the whole distribution. The second contribution is an alternative scoring function at the level of the individual tree estimator, in which we replace the depth-based scoring of the Isolation Forest with one based on hyper-volumes associated to an isolation tree's leaf nodes. 

We motivate the use of both of these methods on generated data and also evaluate them on 34 datasets from the recent and exhaustive ``ADBench'' benchmark, finding significant improvement over the standard isolation forest for both variants on some datasets and improvement on average across all datasets for one of the two variants. The code to reproduce our results is made available as part of the submission.

\end{abstract}

\section{Introduction}

\begin{figure}[t]
    \begin{center}
        \includegraphics[width=55ex]{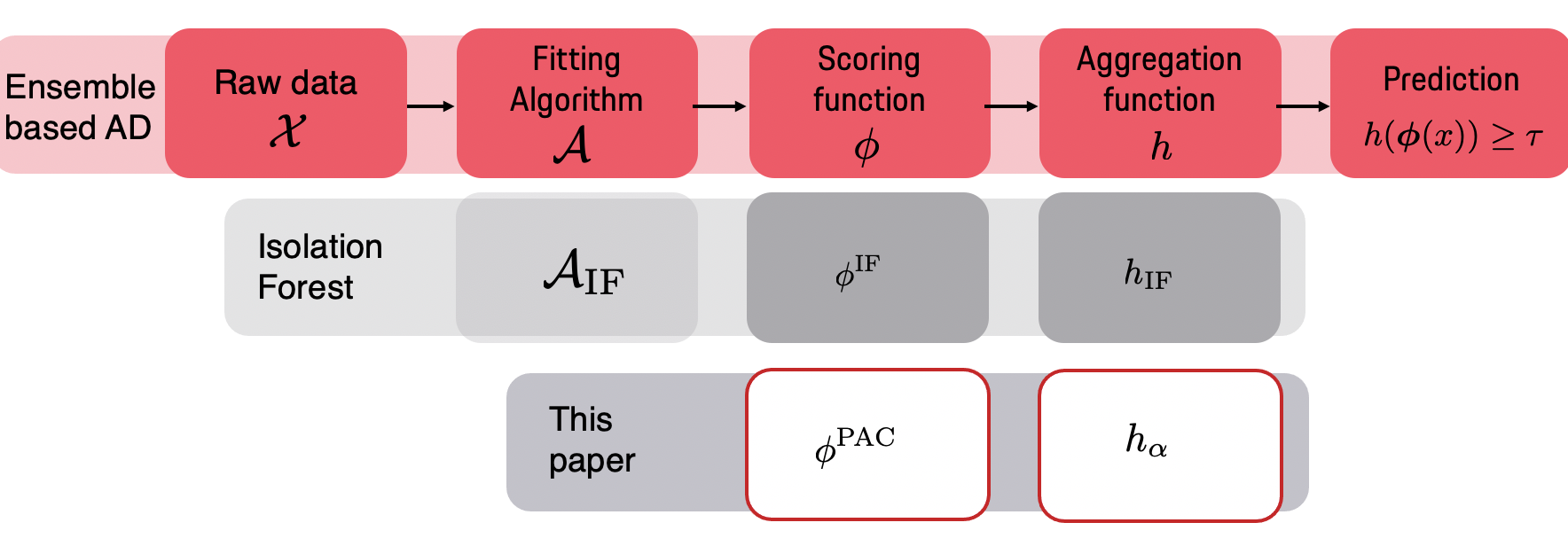}
        \caption{Overview of our framework. The top red boxes indicate the flowchart of an ensemble based anomaly detection. The grey boxes underneath show the specifics of the Isolation Forest. At the bottom in white with red border are the contributions of this paper visualized.}
        \label{fig:AnomalyDetectionOverview}
    \end{center}
\end{figure}

Isolation Forest ($\IF$) \cite{OriginalIsolationForest-short, OriginalIsolationForest-long} is one of the most commonly used machine learning methods for anomaly detection, due to its combination of short fitting and evaluation times with high performance across a variety of different types of anomalies. Abstractly, we can view IF as an ensemble-based anomaly detection method, in which a given (and in general unlabeled) training data set $\mathcal{X} \subset \mathbb{R}^d$ is used to fit an ensemble $(E_i)_{i=1}^n$ of $n$ estimators $E_i$ according to some fitting algorithm that we denote as $\mathcal{A}$ (and that in general would be non-deterministic). This ensemble is then used to classify any given point $x \in \mathbb{R}^d$ as outlier or inlier based on the following two steps: \\
\noindent
First, an \emph{anomaly score} is obtained for each estimator based on a given \emph{per-estimator scoring} function $\phi$ such that $\phi(E_i, x) \in \mathbb{R}$ is the \emph{anomaly score} given to $x$ by the $i$-th estimator. \\
\noindent
Second, the individual scores of the estimators are then aggregated into a final score via an \emph{aggregation function} $h: \mathbb{R}^n \to \mathbb{R}$ such that the IF algorithm classifies $x$ as an anomaly if 
$$
h(\boldsymbol{\phi}(x)) \geq \tau,
$$
where $\tau \in \mathbb{R}$ is some threshold that is specified as part of the input and $\boldsymbol{\phi} = (\phi_1(x), \phi_2(x), \dots, \phi_n(x))$ with $\phi_i(x) := \phi(E_i, x)$. See Fig.\ref{fig:AnomalyDetectionOverview} for an overview of this process.\footnote{
In general, the co-domains of both of these functions could be higher-dimensional, as could be $\tau$, with an appropriate ordering. However, for the sake of simplicity, we limit our presentation to real-valued scoring and aggregation functions only.
}

We can hence specify any ensemble-based anomaly detection method by a tuple $(n, \mathcal{A}, \phi, h, \tau)$. In the specific case of the IF, the fitting algorithm $\mathcal{A}_\IF$ fits a random forest, in which each tree is grown randomly until either a leaf node contains only a single point in a subsampling set $\mathcal{Y} \subseteq \mathcal{X}$ or a maximal depth is reached. The per-estimator scoring function $\phi^\IF$ is then defined such that
$$
\phi^{\mathrm{IF}}(E_i, x) = \frac{d_i(x)}{c_{|\mathcal{Y}|}},
$$ 
where $d_i(x)$ equals the depth of the leaf node associated to point $x$ and $c_{|\mathcal{Y}|}$ is a constant that only depends on the size of the subsampling set used to fit the trees. Finally, the aggregation function $h$ is defined as\footnote{The authors in \cite{OriginalIsolationForest-short} refer to $h_\IF$ as the ``scoring function'', however, given the distinction between the scoring and the aggregation step that we make in this paper we felt the need to move away from that terminology} 
$$
h_\IF(\mathbf{x}) = 2^{- \frac{\sum_{i=1}^n x_i}{n}}, 
$$
that is, the aggregate score is simply the sample mean of the per-estimator scores combined with an outer ``normalization'' function that ensures the score lies in the interval $[0,1]$, anomalies correspond to points with higher anomaly score, and the points are more evenly distributed than under the logarithmic scaling of $\phi^\IF$. In terms of the above tuple notation, IF is then specified as 
$$
\IF \equiv (n, \mathcal{A}_\IF, \phi^\IF, h_\IF, \tau),
$$
where we leave the specification of $n$ and $\tau$ implicit.

\subsection{Contributions}

In this paper, we contribute two logically independent variants of the IF method. In particular, while we don't touch the fitting routing $\mathcal{A}_\IF$ (as the core of the IF algorithm), we study a different \emph{per-estimator scoring function} $\phi^\PAC$ that is inspired by computational learning theory and also introduce a parametrized family of aggregation functions $h_\alpha$ that have $h_\IF$ as a special case. We motivate the use of these variants for anomaly detection by evaluating them both on generated toy datasets, to exemplify their strengths compared to the ``plain vanilla'' $\IF$, as well as on 34 datasets taken from the recent and exhaustive ADBench benchmark \cite{han2022adbench}. We find that our the variant based on the family $h_\alpha$ slightly outperforms the standard $\IF$ on average, while the variant based on the scoring function $\phi_\PAC$ performs worse on average but outperforms $\IF$ and all other unsupervised anomaly detection methods benchmarked in \cite{han2022adbench} on 6 out of 34 datasets.

\subsection{Related Work}

\subsubsection{Generalised Isolation Forest}
The original Isolation Forest is considered a state-of-the-art anomaly detection algorithm. However, in some cases the limits of this algorithm are reflected in a poor anomaly detection performance. Among the disadvantages is the random choice of the split value used in the construction of the trees. The performance of this algorithm also suffers when it comes to dealing with high-dimensional data or clusters of anomalies.
There exist several IF generalisations which tackle the weaknesses of the original Isolation Forest paper. 
In Deep Isolation Forest~\cite{Xu_2023} a deep neural network architecture is used, in the HDoutliers algorithm~\cite{talagala2019anomaly}, extreme value theory is used to calculate a more suitable threshold for the anomalies and in PIDForest~\cite{gopalan2019pidforest} a feature selection mechanism is introduced to identify the most relevant features to improve the performance on high dimensional data, which is known as the curse of dimensionality. In Extended Isolation Forest~\cite{Hariri_2021} and Generalized IF~\cite{Lesouple2021} feature subsetting and categorical feature handling mechanisms, and different distance measures were added to tackle high dimensional datasets and mixed data types. In SCIForest~\cite{Liu2010OnDC} a sliding window approach is introduced to process stream data. In~\cite{Mensi2021} different aggregation function are used to combine the tree scores. In~\cite{Tokovarov2022} a probabilistic mechanism is used that allows to choose better split values to detect anomalies hidden between clusters more effectively. 

\subsubsection{Computational learning for anomaly detection}

Many recent works around PAC (Probably Approximately
Correct) and anomaly detection have tried to find guarantees for semi-supervised  \cite{Li_2022} and unsupervised learning \cite{PACSampleComplexityAnomalyDetection}. This research direction is motivated by the need for guarantees in order to bring learning tasks into (safety critical) real world applications like autonomous driving \cite{pmlr-v205-farid23a}, manufacturing \cite{niggemann2013learning} and internet of medical things \cite{Ji2023IncrementalAD}. This work \cite{PACSampleComplexityAnomalyDetection} investigates the sample complexity, in the PAC context, of unsupervised anomaly detection. We re-implemented and generalized their PAC motivated algorithm, since their experimental results where not reproducible. This generalization is practically motivated and might not inherit their theoretical proofs. 

\subsection{Code}

We make all code required to reproduce the results available.\footnote{\url{https://github.com/porscheofficial/distribution_and_volume_based_isolation_forest}}

\section{Method}

In this section, we introduce the two main contributions of this work, a novel per-estimator score function and a novel aggregation function for ensemble-based anomaly detection methods. As they are logically independent, we introduce them in separate subsections. In each case, we first define them formally and then motivate their usefulness in the context of anomaly detection, using generated datasets. In section \ref{sec:experiments}, we'll apply these methods to actual benchmark datasets.

\subsection{Distribution-based aggregation functions}

As discussed in the introduction, an important part of an ensemble-based anomaly detection method is the question how to aggregate the individual per-estimator scores $\phi_i(x)$ into an aggregate score, in order to classify a point. As we've seen, in IF (a function of) the sample mean is used is to aggregate. This is a natural and intuitive choice, however, there can be situations, in which a good estimator will take into consideration features of the vector of estimator scores $\boldsymbol{\phi}(x)$ other than its sample mean. 

For example, consider a dataset $\mathcal{X} \subset \mathbb{R}^d$ that contains $N-1 \gg 1$ inlier points drawn at random from the $d$-dimensional unit cube and a single outlier point $\tilde{x} = (1.05, 0.5, \dots, 0.5)$ that lies just slightly outside of this cube along a single dimension. In this setup, it is possible to separate this point using only a single decision node in a random tree, however, sampling such a tree is statistically relatively unlikely and, as such, most estimators will contain $\tilde{x}$ in a leaf node at a depth similar to the inlier points. Yet, for large $N$, it is, on average, significantly \emph{more} likely for a random tree to separate $\tilde{x}$ using very few decision nodes than it is for inlier points. As such, we intuitively expect that for sufficiently large $n$, there would be significantly more estimators that isolate $\tilde{x}$ at low depth than there are estimators that isolate any inlier points at that depth. For this reason, we would expect an aggregation function that places disproportionate weight onto high per-estimator scores to lead to a classifier that outperforms one that only considers the sample mean across all estimators.

An extreme version of this idea would be to place all the weight onto the single estimator with the \emph{worst} anomaly score, however, such an extreme weighing mechanism would surely not lead to a robust classifier and more often than not ``throw away the baby with the bath water''. Instead, in this paper, we introduce a \emph{family} of aggregation functions, parametrized by a single parameter $\alpha$, that lets users ``tune'' the sensitivity of the aggregation step to estimators with below-average anomaly scores. Specifically, define the functions $f_\alpha: \mathbb{R}^d_{\geq 0} \to \mathbb{R}$, for $\alpha \in [0,1) \cup (1,\infty)$, via the mapping

$$
f_\alpha(\mathbf{x}) = n^{\frac{1}{\alpha -1}}\left(\sum_{i=1}^n x_i^{1-\alpha}\right)^{\frac{1}{1-\alpha}}
$$

and $f_\beta = \lim_{\alpha \to \beta} f_\alpha$ for $\beta \in \{1, \infty \}$. This leads to the corresponding aggregation functions for $\alpha \geq 0$,
$$
h_\alpha(\mathbf{x}) = 2^{-f_\alpha(\mathbf{x})}
$$

In Appendix \ref{app:renyi}, we use a connection of these functions to the \emph{R\'{e}nyi divergences} from information-theory to show that they satisfy the following properties: satisfy the following properties:
\begin{enumerate}
\item $h_0 = h_\IF$
\item $h_\infty(\mathbf{x}) = \max( \mathbf{x})$
\item $\alpha \geq \alpha' \Rightarrow h_\alpha(\mathbf{x}) \geq f_{\alpha'}(\mathbf{x}), \quad \forall \: \mathbf{x} \in \mathbb{R}^d_{\geq 0}, \alpha \geq 0$
\end{enumerate}

In other words, the aggregation functions $h_\alpha$ functions are monotonically increasing in $\alpha$ and interpolate between the standard $IF$ aggregation function on the one end ($\alpha = 0$) and the maximum on the other end ($\alpha = \infty$). In this sense, the above aggregation functions introduce a strict generalisation of the original IF algorithm and we can write
$$
\IF_\alpha \equiv (n, \mathcal{A}_\IF, \phi^\IF, h_\alpha, \tau),
$$
with $\IF \equiv \IF_0$. 

The monotonicity property of the aggregation functions proves to be very convenient for thinking about the effect of changing the value of $\alpha$: All else being equal, increasing $\alpha$ can only \emph{increase} the number of points a classifier considers anomalous, while conversely reducing it can only \emph{decrease} this number. It is for this reason that we call $\alpha$ the ``sensitivity'' of the classifier. As a corollary, for any $n, \mathcal{A}, \phi, \tau$, and $\alpha > \alpha'$, the anomaly detector
$$
AD_\alpha \equiv (n, \mathcal{A}, \phi, h_\alpha, \tau)
$$
will have a false negative rate smaller or equal to that of the corresponding detector $AD_{\alpha'}$ (at the cost of a potentially larger false positive rate).\footnote{As changing the threshold also has the effect of changing the predictions in only one direction, it may seem as if the effect of changing $\alpha$ can similarly be obtained by changing $\tau$. That is, the reader might be under the impression that, continuing the example in the main text, for any value $\alpha'$, there would in general exist a value $\tau'$ such that the predictions of the classifier $(n, \mathcal{A}, \phi, h_\alpha, \tau')$ coincide with those of $AD_{\alpha'}$. This is, however, not the case, as changing $\alpha$ can lead to a different \emph{relative ordering} of data points wrt their aggregated score, while this is impossible by changing $\tau$.}

\begin{figure}[!htbp]
\begin{center}
\subfloat{\includegraphics[height=20ex]{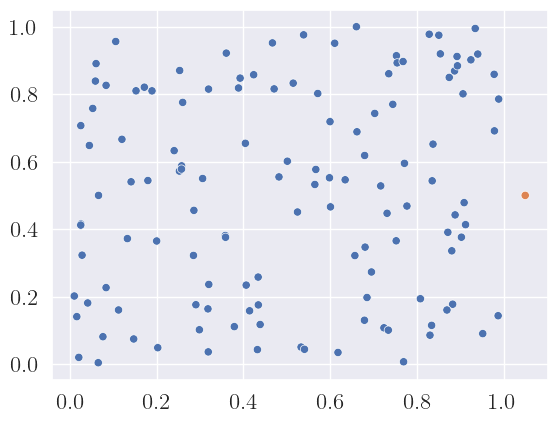}}
\subfloat{\includegraphics[height=20ex]{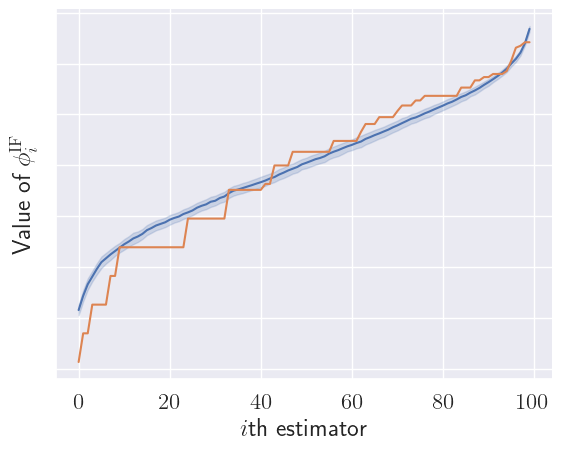}}\\
\subfloat{\includegraphics[height=20ex]{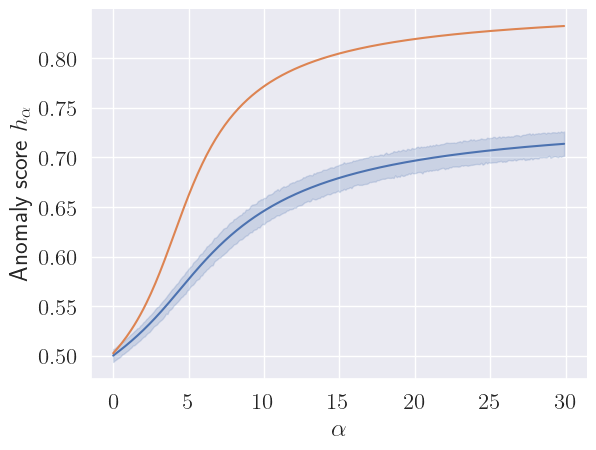}}
\subfloat{\includegraphics[height=20ex]{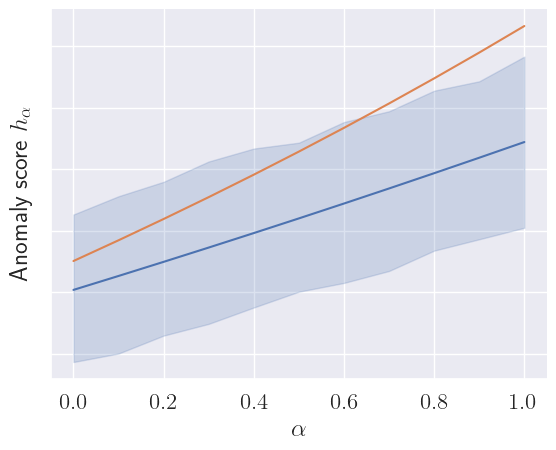}}
\end{center}
\caption{Motivating the use of distribution-based aggregation functions: We sample a dataset consisting of $N=128$ points with $127$ inliers (blue) uniformly sampled from a $d$-dimensional unit cube and a single anomalous point (orange) slightly outside along a single dimension \emph{(Top left)}. Fitting an isolation forest with $n=100$ estimators to such a dataset leads to a distribution of (sorted) per-estimator scores in which the anomaly receives a ``tail'' of significantly worse scores than most inliers \emph{(Top right)}. As a result, we can separate the outlier from the inliers by using an aggregation function with sufficiently large $\alpha$ \emph{(Bottom left)}. In particular, zooming into this curve for $\alpha \leq 1$, we see that using the sample mean ($\alpha = 0$) leads to an outlier anomaly score better than many inliers (blue shaded regions correspond to the 95th percentile of inlier scores) \emph{(Bottom right)}. Running the above experiment a hundred times 
 for $d=10$ leads to average AUCROC score of $0.78$ for the $\IF$ algorithm, as opposed to $0.98$ when using $\IF_\infty$.}
\label{fig:renyi_experiments}
\end{figure}

We believe that the above family of aggregation functions introduces a practically useful degree of freedom for designing an anomaly detection method. Indeed, in Fig.~\ref{fig:renyi_experiments} we describe and analyse experiments, based on the motivating toy example above, that show a significant increase in performance for $IF_\alpha$ when increasing $\alpha$.

\subsection{Hypervolume-based scoring function}
The other important part of an
ensemble-based anomaly detection method is the question how
to choose the per-estimator scores $\phi_{i}(x)$ for a given data point $x$. 

The standard $\IF$ algorithm uses random trees as estimators. As such, the leaf nodes $(L^{(i)}_j)_j$ of the $i$th estimator $E_i$ by construction partition the space $\mathbb{R}^d$ into a set of axis-aligned hyper-rectangles, one for each leaf node. The boundaries of these rectangles simply correspond to the threshold tests at the decision nodes that lead up to a leaf node. With slight abuse of notation, in the following we will use $L^{(i)}_j$ to denote both this rectangle as well as the leaf node. Moreover, we will use $L^{(i)}(x)$ to denote that hyperrectangle in the partition corresponding to $E_i$ that contains the data point $x$.

Instead of the classical IF scoring based on the depths of the leaf nodes, we propose a scoring function motivated by the Probably Approximately Correct (PAC) framework. The PAC-RPAD framework is a method developed in \cite{PACSampleComplexityAnomalyDetection} to show polynomial sample complexity for learning an unsupervised anomaly detection method with proven guarantees. Inspired by this framework, we define the per-estimator scoring function 
\begin{equation}
\phi^{PAC}(E_i, x) = \frac{|L^{(i)}(x)|}{|\mathcal{Y}|} \cdot \frac{V(\mathcal{Y})}{V(L^{(i)}(x))},
\label{eq:pac}
\end{equation}
where $|L^{(i)}_j|$ denotes the number of points of the subsampling set $\mathcal{Y}$ that are contained in $L^{(i)}_j$, $V(L)$ denotes the volume of the hyperrectangle $L$ and $V(\mathcal{Y})$ denotes some bounding volume that contains all points in $\mathcal{Y}$ but whose total size is a hyperparameter of the algorithm.

To understand the motivation behind this scoring function, note that the first of the two factors in the RHS of \eqref{eq:pac} constitutes a density estimate of the likelihood of sampling a given hyper-rectangle when randomly sampling a point in the subsampling set, while the second factor measures the relative volume of this rectangle, which coincides with the probability of sampling this rectangle when sampling a point in the bounding volume uniformly at random. As such, points with bad (i.e. low) anomaly score are those that have density significantly below a uniform. See \cite{PACSampleComplexityAnomalyDetection} for a detailed discussion. The heuristic behind this choice of scoring function is that outliers $x$ are expected to be located inside hyperrectangles which occur with small frequency.

Using the above notation and the concepts from the previous section, we then introduce the family of anomaly detection methods
$$
\PAC_\alpha \equiv (n, \mathcal{A}_\IF, \phi^\PAC, h_\alpha, \tau).
$$

\begin{figure}[!tbp]
\begin{center}
\subfloat{\includegraphics[height=22ex]{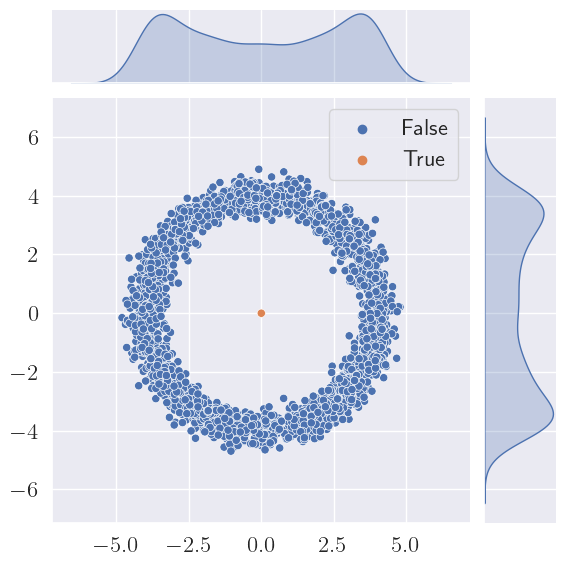}} \quad
\subfloat{\includegraphics[height=22ex]{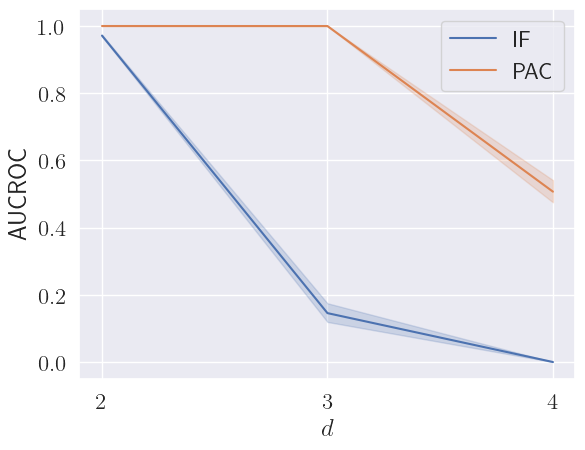}}\\
\end{center}
\caption{Motivating the use of a hypervolume-based scoring function: We sample a dataset consisting of $N=127$ points uniformly from the surface of a $d-1$ dimensional sphere (with some additional Gaussian noise) and place a single anomaly at the origin. The anomaly is easy to separate in the full-dimensional space but ``screened'' in any lower-dimensional projection of the data \emph{(Left)}. We fit these data to $\IF_0$ and $\PAC_0$ respectively. Running this experiment a hundred times shows a strong and robust separation between the $\mathrm{AUCROC}$s the two algorithms, especially for $d=3$. The reason for the collapse of both algorithms for increasing $d$ we assume to be an exponentially small likelihood of sampling random trees whose decision nodes imply full-dimensional hyperrectangles \emph{(Right)}.}
\label{fig:volume_experiments}
\end{figure}

Just as in the case of the aggregation functions $h_\alpha$, an obvious question is under which circumstances we expect hypervolume-based scoring to outperform depth-based scoring. While we do not have a systematic answer to this question, in the following we want to present a simple example in which we found a significant difference in their performance: Consider a dataset $\mathcal{X}$ that consists of $N-1 \gg 1$ inlier points randomly distributed on the $d-1$-dimensional unit sphere and a single outlier point at the origin. In this case, on average, the largest, in volume, axis-aligned hyper-rectangle that contains only the outlier is significantly larger than than that of every inlier point. While a similar statement holds true for the average depth, it is not nearly as pronounced as for the volume. This separation between inliers and outlier increases in $d$, hence we expect $\PAC_\alpha$ to outperform $\IF_\alpha$ as the dimension increases. This expectation is confirmed by experiments we ran for the above setup and that show a breakdown of the $\mathrm{AUCROC}$ score of $IF_0$ for $d>2$. See Fig.~\ref{fig:volume_experiments} for details.

\section{Experiments}
\label{sec:experiments}
\subsection{Setup}

We used the recently published ADBench  \cite{han2022adbench} suite of benchmark datasets to compare the performance of the anomaly detection methods $\IF_\alpha$ and $\PAC_\alpha$ introduced in this paper with that of the standard isolation forest and other unsupervised anomaly detection methods, for the values $\alpha \in \{0,0.5,1,2,\infty\}$. As common, we used the $\mathrm{AUCROC}$ metric as the score metric, using the values as stated in \cite{han2022adbench}. We ran the experiments on 34 datasets, using the code provided in \cite{han2022adbench} and using a custom implementation of the above methods. For the isolation forest and its variants, we used 100 estimators and evaluated datasets for which an AUCROC value was computed. This means that we ignored datasets for which the AUCROC value was "None" in the ADBench paper. \footnote{For reasons not entirely clear to us, we could not reproduce the results for the isolation forest that are stated in \cite{han2022adbench}, even when using the authors' provided scripts and code without any modifications. Indeed, we found an average of 0.06 $\mathrm{AUCROC}$ points across 34 datasets. While this means that the comparison between the IF variants of this paper and the other algorithms should be interpreted with care, it does not affect the comparison between those variants.}

\subsection{Results}

\begin{figure}[tb]
\centering
\includegraphics[scale=0.3]{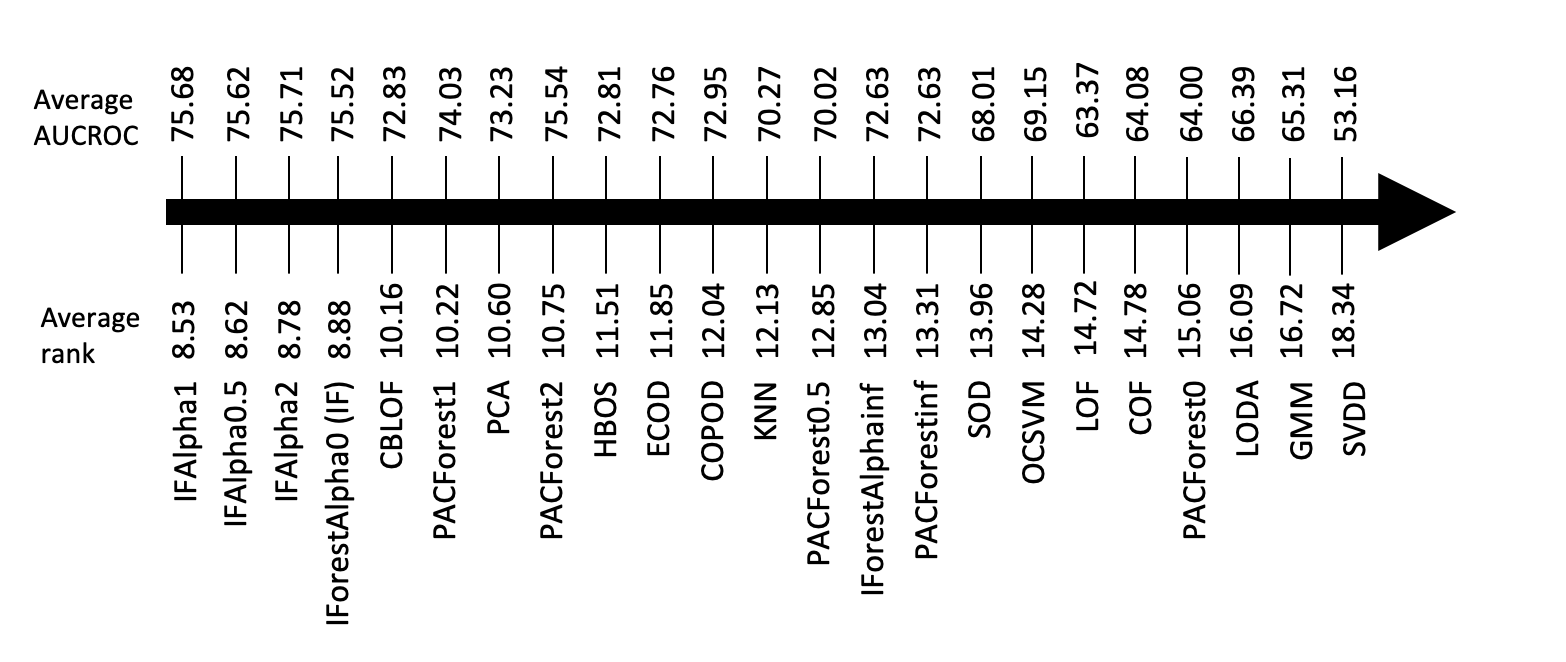}
\caption{Ranking of the algorithms according to performance: Following \cite{han2022adbench}, to generate the ranking, we ranked all algorithms on each dataset by $\mathrm{AUCROC}$ and then averaged the rank over datasets. For completeness we also state to the average $\mathrm{AUCROC}$ across the considered datasets.}
\label{fig:rank_table}
\end{figure}

\begin{figure}[h]
\caption{Boxplot of AUCROC IF variants and SOTA algorithms}
\centering
\includegraphics[scale=0.4]{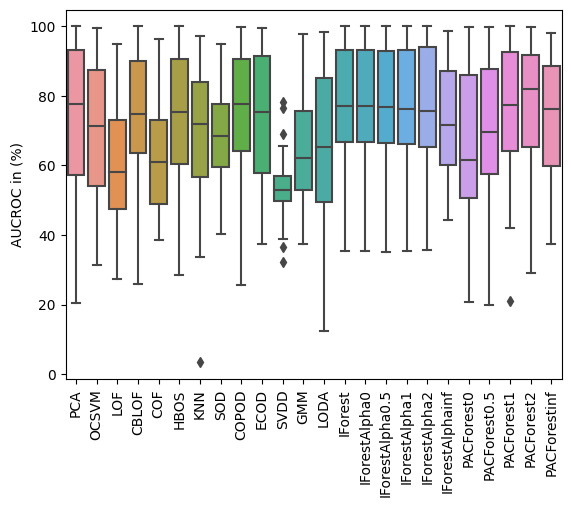}
\label{fig:boxplot_aucroc_sota_and_IF_variants}
\end{figure}

We obtained the following results:

\begin{enumerate}
    \item On the ADBench datasets, $\IF_\alpha$ slightly outperforms $\IF$ on average across datasets, for $\alpha \in \{0.5,1,2\}$, while $\PAC_\alpha$ on average performs significantly poorer (see Fig.~\ref{fig:rank_table}).
    \item Consistent with the results of \cite{han2022adbench}, the variance across datasets in the performance of unsupervised anomaly detection methods is high and the algorithms defined herein are no exception to this (see Fig.~\ref{fig:boxplot_aucroc_sota_and_IF_variants}).
    \item While poor on average, there are datasets in which $\PAC_\alpha$ anomaly detection algorithms significantly outperform their counterpart $\IF_\alpha$ and state-of-the-art methods. To illustrate this, Tab.~\ref{Tab:AUCROC_SOTA_and_IF_variants_small} in the appendix provides an overview of the performances of $\PAC_\alpha$ against $\IF_\alpha$ and the best of the other comparison algorithms, for a selected subset of the datasets.
\end{enumerate}

\subsection{Discussion}

The findings confirm the verdict from \cite{han2022adbench} that there is not one anomaly detection method that suits all purposes, and the anomaly detection methods discussed in this paper are no exception. At the same time, we see that the introduction of distribution-based scoring has an overall positive, albeit mild, impact on the performance of $\IF$, when $\alpha$ is relatively small, while performance significantly decreases as $\alpha \to \infty$. This confirms the intuition that \emph{some} sensitivity for outlier votes across estimators helps with anomaly detection, while too much of it does not. 

Regarding our volume-based variant of $\IF$, the success hinges very much on the dataset: On some datasets, we found a significant improvement over $\IF$ and other unsupervised algorithms, however, on most datasets performance was poor. Unfortunately, at this point we lack a systematic understanding of the criteria that determine whether $\PAC_\alpha$ is a suitable method for a given dataset, in the absence of ground truth data.

\section{Conclusion and Future Work}

To conclude, in this note we introduced, motivated and empirically tested two families of ensemble-based anomaly detection methods, $\IF_\alpha$ and $\PAC_\alpha$, both variants of the successful Isolation Forest algorithm. We evaluated them on 34 datasets from the recent and exhaustive ``ADBench'' benchmark, finding significant improvement over the standard isolation forest for both variants on some datasets and improvement on average across all datasets for $\IF_\alpha$. We briefly discussed these results.

For future work, we believe that it could be beneficial to apply the distribution-based aggregation functions $h_\alpha$ also to other ensemble-based anomaly detection methods, such as LODA \cite{LODA}, or even ensemble-based methods in machine learning beyond anomaly detection. We also believe that it would be valuable to research the possibility of an ``auto-tuning'' mechanism for $\alpha$, that is, some way of tuning $\alpha$ based on an unlabeled training dataset.
Finally, regarding $\PAC_\alpha$, we believe that it would be interesting to better understand the types of anomaly datasets on which hypervolume-based scoring is a better choice than depth-based scoring.

\section*{Acknowledgment}
We thank Dr. Mahdi Manesh, Regina Kirschner, and the Porsche PANAMERA project for valuable discussions, as well as Prof. Jean-Pierre Seifert and Niklas Pirnay for discussions during the early phase of this project.

\bibliographystyle{plain}
\bibliography{bib.bib}

\begin{thebibliography}{10}

\bibitem{pmlr-v205-farid23a}
Alec Farid, Sushant Veer, Boris Ivanovic, Karen Leung, and Marco Pavone.
\newblock Task-relevant failure detection for trajectory predictors in
  autonomous vehicles.
\newblock In Karen Liu, Dana Kulic, and Jeff Ichnowski, editors, {\em
  Proceedings of The 6th Conference on Robot Learning}, volume 205 of {\em
  Proceedings of Machine Learning Research}, pages 1959--1969. PMLR, 14--18 Dec
  2023.

\bibitem{gopalan2019pidforest}
Parikshit Gopalan, Vatsal Sharan, and Udi Wieder.
\newblock Pidforest: Anomaly detection via partial identification.
\newblock 2019.

\bibitem{han2022adbench}
Songqiao Han, Xiyang Hu, Hailiang Huang, Mingqi Jiang, and Yue Zhao.
\newblock Adbench: Anomaly detection benchmark.
\newblock 2022.

\bibitem{Hariri_2021}
Sahand Hariri, Matias~Carrasco Kind, and Robert~J. Brunner.
\newblock Extended isolation forest.
\newblock {\em {IEEE} Transactions on Knowledge and Data Engineering},
  33(4):1479--1489, apr 2021.

\bibitem{Ji2023IncrementalAD}
Xiayan Ji, Hyonyoung Choi, Oleg Sokolsky, and Insup Lee.
\newblock Incremental anomaly detection with guarantee in the internet of
  medical things.
\newblock {\em Proceedings of the 8th ACM/IEEE Conference on Internet of Things
  Design and Implementation}, 2023.

\bibitem{Lesouple2021}
Julien Lesouple, C{\'{e}}dric Baudoin, Marc Spigai, and Jean-Yves Tourneret.
\newblock Generalized isolation forest for anomaly detection.
\newblock {\em Pattern Recognition Letters}, 149:109--119, September 2021.

\bibitem{Li_2022}
Shuo Li, Xiayan Ji, Edgar Dobriban, Oleg Sokolsky, and Insup Lee.
\newblock {PAC}-wrap.
\newblock In {\em Proceedings of the 28th {ACM} {SIGKDD} Conference on
  Knowledge Discovery and Data Mining}. {ACM}, aug 2022.

\bibitem{OriginalIsolationForest-short}
Fei~Tony Liu, Kai~Ming Ting, and Zhi-Hua Zhou.
\newblock Isolation forest.
\newblock pages 413--422, 2008.

\bibitem{Liu2010OnDC}
Fei~Tony Liu, Kai~Ming Ting, and Zhi-Hua Zhou.
\newblock On detecting clustered anomalies using sciforest.
\newblock 2010.

\bibitem{OriginalIsolationForest-long}
Fei~Tony Liu, Kai~Ming Ting, and Zhi-Hua Zhou.
\newblock Isolation-based anomaly detection.
\newblock {\em ACM Trans. Knowl. Discov. Data}, 6:3:1--3:39, 2012.

\bibitem{Mensi2021}
Antonella Mensi and Manuele Bicego.
\newblock Enhanced anomaly scores for isolation forests.
\newblock {\em Pattern Recognition}, 120:108115, December 2021.

\bibitem{niggemann2013learning}
Oliver Niggemann, Asmir Vodencarevic, Alexander Maier, Stefan Windmann, and
  Hans~Kleine B{\"u}ning.
\newblock A learning anomaly detection algorithm for hybrid manufacturing
  systems.
\newblock In {\em The 24th International Workshop on Principles of Diagnosis
  (DX-2013), Jerusalem, Israel}, 2013.

\bibitem{LODA}
Tom{\'a}{\v s} Pevn{\'y}.
\newblock Loda: Lightweight on-line detector of anomalies.
\newblock {\em Machine Learning}, 102(2):275--304, 2016.

\bibitem{renyi1961measures}
Alfr{\'e}d R{\'e}nyi.
\newblock On measures of entropy and information.
\newblock In {\em Proceedings of the Fourth Berkeley Symposium on Mathematical
  Statistics and Probability, Volume 1: Contributions to the Theory of
  Statistics}, volume~4, pages 547--562. University of California Press, 1961.

\bibitem{PACSampleComplexityAnomalyDetection}
Md~Amran Siddiqui, Alan Fern, Thomas~G. Dietterich, and S.~Das.
\newblock Finite sample complexity of rare pattern anomaly detection.
\newblock 2016.

\bibitem{talagala2019anomaly}
Priyanga~Dilini Talagala, Rob~J. Hyndman, and Kate Smith-Miles.
\newblock Anomaly detection in high dimensional data.
\newblock 2019.

\bibitem{Tokovarov2022}
Mikhail Tokovarov and Pawe{\l} Karczmarek.
\newblock A probabilistic generalization of isolation forest.
\newblock {\em Information Sciences}, 584:433--449, January 2022.

\bibitem{Renyi_properties}
Tim van Erven and Peter Harremos.
\newblock Rényi divergence and kullback-leibler divergence.
\newblock {\em IEEE Transactions on Information Theory}, 60(7):3797--3820,
  2014.

\bibitem{Xu_2023}
Hongzuo Xu, Guansong Pang, Yijie Wang, and Yongjun Wang.
\newblock Deep isolation forest for anomaly detection.
\newblock {\em {IEEE} Transactions on Knowledge and Data Engineering}, pages
  1--14, 2023.

\end{thebibliography}

\appendix

\subsection{Properties of $f_\alpha$}
\label{app:renyi}
To clarify the connection of the functions $f_\alpha$ to information theory and show the claimed properties, let us first define the $\alpha$-R\'enyi divergence for $\alpha \in (0,1) \cup (1,\infty)$, and $p, q \in \mathbb{R}^n_{\geq 0}$ as
$$
R_\alpha(p \| q) = \frac{1}{\alpha - 1} \ln\sum_{i=1}^d p_i^\alpha q_i^{1-\alpha}.
$$
with $R_\beta(p\|q) = \lim_{\alpha \to \beta} R_\alpha(p\|q)$ for $\beta \in \{0,1,\infty\}$. The R\'enyi divergences, introduced in \cite{renyi1961measures}, generalize the Kullback-Leibler divergence, or relative entropy (which corresponds to $\alpha = 1$) and have a variety of use cases in information \cite{renyi1961measures, Renyi_properties}. Here, we link them to the aggregation functions introduced in the main text via the simple identity 
$$
f_\alpha(\mathbf{x}) = \exp\left(-R_\alpha\left(\frac{\mathbf{1}}{n} \| \frac{\mathbf{x}}{n}\right)\right),
$$
where $\mathbf{1}$ is the vector of ones. The claimed properties of $f_\alpha$ are then a direct consequence of the properties of the Renyi divergences, see e.g. Theorems 3, 4 and 6 in \cite{Renyi_properties}.

\clearpage
\begin{table}[b]
\begin{minipage}{\textwidth}
\centering
\caption{Detailed results for the $\mathrm{AUCROC}$ for SOTA algorithms and IF variants on selected datasets. The highest $\mathrm{AUCROC}$ is highlighted in bold. The best value of $\alpha$ and best SOTA algorithm are given in parentheses behind the $\mathrm{AUCROC}$ value.}
\resizebox{300pt}{!}{%
\begin{tabular}{llll}
Dataset &         best $\IF_\alpha$ &        best $\PAC_\alpha$ & best SOTA             \\ \\
annthyroid &    82.12 (0) &  \textbf{91.47} (0.5) &     82.12 (IForest) \\
cover      &  \textbf{96.58} ($\infty$) &  92.57 ($\infty$) &         93.73 (PCA) \\
landsat    &    48.79 (1) &    \textbf{66.24} (0) &        63.61 (SVDD) \\
thyroid    &  97.84 (0.5) &    \textbf{98.85} (0) &     97.83 (IForest) \\
Waveform   &    69.49 (1) &    \textbf{86.88} (0) &       75.03 (COPOD) \\
wine       &   76.4 (0.5) &  \textbf{92.59} (0.5) &        91.36 (HBOS) \\
WPBC       &    52.66 (0) &  \textbf{63.63} (0.5) &     52.66 (IForest) \\
\end{tabular}%
}
\label{Tab:AUCROC_SOTA_and_IF_variants_small}
\end{minipage}
\end{table}

\end{document}